\documentclass[sigconf]{acmart}

\AtBeginDocument{%
  }
\setcopyright{acmlicensed}
\copyrightyear{2024}
\acmYear{2024}

\acmConference[Accepted at the CONSEQUENCES ‘24 workshop, co-located with ACM RecSys ‘24]{October, 2024}{Bari, Italy}
  
\usepackage{graphicx}   
\usepackage{caption}    
\usepackage{subcaption} 
\usepackage[utf8]{inputenc} 
\usepackage[T1]{fontenc}    
\usepackage{hyperref}       
\usepackage{url}            
\usepackage{booktabs}       
\usepackage{nicefrac}       
\usepackage{microtype}      
\usepackage{xcolor}         
\usepackage{siunitx}        
\usepackage{graphicx}       
\usepackage{algorithm}      
\usepackage{algorithmic}    
\usepackage{amsmath}        
\usepackage{amsthm}         
\usepackage{multirow}

\newtheorem{proposition}{Proposition}

\title{Causal Feature Selection Method for Contextual Multi-Armed Bandits in Recommender System}

\author{Zhenyu Zhao}
\email{zzhao@roblox.com}
\affiliation{%
  \institution{Roblox}
  \city{San Mateo}
  \state{California}
  \country{USA}
}

\author{Yexi Jiang}
\email{hjiang@roblox.com}
\affiliation{%
  \institution{Roblox}
  \city{San Mateo}
  \state{California}
  \country{USA}
}

\begin{document}

\begin{abstract}
Effective feature selection is essential for optimizing contextual multi-armed bandits (CMABs) in large-scale online systems, where suboptimal features can degrade rewards, interpretability, and efficiency. Traditional feature selection often prioritizes outcome correlation, neglecting the crucial role of heterogeneous treatment effects (HTE) across arms in CMAB decision-making. This paper introduces two novel, model-free filter methods, Heterogeneous Incremental Effect (HIE) and Heterogeneous Distribution Divergence (HDD), specifically designed to identify features driving HTE. HIE quantifies a feature's value based on its ability to induce changes in the optimal arm, while HDD measures its impact on reward distribution divergence across arms. These methods are computationally efficient, robust to model mis-specification, and adaptable to various feature types, making them suitable for rapid screening in dynamic environments where retraining complex models is infeasible. We validate HIE and HDD on synthetic data with known ground truth and in a large-scale commercial recommender system, demonstrating their consistent ability to identify influential HTE features and thereby enhance CMAB performance.
\end{abstract}

\maketitle


\section{Introduction}
Multi-armed bandits (MABs) \cite{lai1985asymptotically, slivkins2019introduction, bouneffouf2024tutorial, heyden2024budgeted} and their contextual counterparts (CMABs) \cite{li2010contextual, tang2014ensemble} are pivotal for adaptive decision-making in dynamic environments like online recommender systems \cite{gangan9survey, kawale2018multiarmed, mcinerney2018explore, yi2023online}. CMABs leverage contextual features to personalize arm selections, aiming to maximize rewards. The efficacy of CMABs, however, critically depends on the quality of these contextual features. Missing influential features can lead to suboptimal policies, while including irrelevant ones increases model complexity, computational cost, and the risk of overfitting.

A central challenge in feature selection for CMABs is that conventional methods, often focused on outcome prediction or correlation \cite{bolon2013review, chandrashekar2014survey, zhao2019maximum,agrawal2021metaheuristic,theng2024feature, lin2022adafs}, inadequately capture the nuances of arm-specific performance. For CMABs, the most valuable features are those that induce \textit{heterogeneous treatment effects (HTE)}, meaning they cause the relative attractiveness of different arms to vary across different contexts \cite{wager2018estimation}. Identifying such HTE-driving features is paramount for effective personalization.

While methods for estimating HTE have advanced significantly in causal inference \cite{kunzel2019metalearners, nie2020quasi, kennedy2020optimal, van2011targeted}, their application as feature selectors in large-scale CMAB systems remains underexplored. Recent work has reduced CMABs to HTE estimation for decision-making \cite{carranza2022flexible}. However, efficient model-free feature selection for multi-arm CMABs - particularly methods avoiding complex policy optimization or restrictive model assumptions - remains an open challenge. Existing CMAB feature importance techniques either rely on model-embedded signals \cite{li2010contextual} or require iterative policy retraining \cite{wang2023policy, foster2019model}, making them computationally prohibitive for large-scale feature screening.

To address these limitations, we propose two novel filter methods for HTE-driven feature selection in CMABs: Heterogeneous Incremental Effect (HIE) and Heterogeneous Distribution Divergence (HDD). HIE quantifies a feature's value through context-specific optimal arm selection gains, while HDD measures its impact on reward distribution divergence across arms. Both methods operate model-free, avoiding mis-specification risks of embedded approaches. They also handle continuous/categorical features and nonlinear HTE patterns. Both methods offer computationally efficient HIE/HDD scores suitable for rapid filtering. Optional bootstrap normalization, which is parallelizable and thus scalable, can further debias these scores and provide p-values.

We demonstrate the effectiveness of HIE and HDD through comprehensive experiments. On synthetic data with diverse, known HTE patterns (Section~\ref{sec:synthetic_evaluation}), our methods consistently outperform traditional feature selectors and show advantages over MAB-reward-based feature ranking, especially in identifying non-linear HTE. In a large-scale deployment within a commercial recommender system (Section~\ref{sec:online_experiments}), we demonstrate the practical utility of HIE and HDD for efficient feature engineering. By first using these methods to screen a vast pool of candidate features-thereby avoiding the prohibitive cost of experimentally testing all of them-we identified a smaller set of high-potential features. Subsequent online A/B testing of CMABs built with these top-ranked features revealed a strong association: features assigned high HIE/HDD scores frequently corresponded to CMABs exhibiting statistically significant deviations from non-contextual behavior at the individual content level. This underscores our methods' real-world applicability for robustly identifying features that enable meaningful contextualization, crucial in scenarios demanding efficient feature selection.

\section{Feature Selection Methods for Heterogeneous Effects in Contextual Multi-Armed Bandits}
In CMAB problems, a feature is considered important if the reward distributions of the arms vary as a function of the feature's values. Specifically, a feature is deemed crucial if it alters the optimal arm selection across different feature values, allowing the contextual MAB to optimize rewards by leveraging contextual information. In this work, we focus on identifying features that are \textit{indicative of heterogeneous treatment effects (HTE)} across arms. While formal causal discovery using techniques like instrumental variables or explicit confounding adjustment is beyond the scope of our model-free filter approach, our methods leverage principles from HTE analysis to quantify how features influence reward distributions and optimal arm selection differently across contexts. This provides a practical, computationally efficient way to screen for features that capture valuable heterogeneity in large-scale systems where full causal modeling may be intractable.

In causal inference, HTE refers to the variation in treatment effects across different subpopulations or feature values \cite{zhao2022feature}. Formally, given a treatment variable $T$ and an outcome $Y$, the individual treatment effect (ITE) for a subject with features $x$ is often defined as $\tau(x) = \mathbb{E}[Y \mid T=1, X=x] - \mathbb{E}[Y \mid T=0, X=x]$. A feature $X$ is important for inducing HTE if $\tau(x)$ varies significantly across different values of $X$. In the CMAB setting, while there isn't always an explicit control group, the concept of HTE remains relevant as the relative differences in rewards across multiple arms depend on feature values.

Given a set of $k$ arms $\mathcal{A} = \{a_1, \dots, a_k\}$ and a reward function $Y(a, x)$ (often binary in our context, e.g., click/no-click), let $P(Y=1|a,x)$ be the probability of receiving a reward for arm $a$ given context $X=x$. The best arm for context $x$ is $a^*(x) = \arg\max_{a \in \mathcal{A}} P(Y=1|a,x)$. A feature $X$ is important if $a^*(x)$ varies across values of $X$. More broadly, a feature can be important if it alters the reward distribution across arms, even without changing the empirically observed best arm, as this can influence exploration-exploitation strategies or indicate an expected change in the best arm. We term such features "HTE features."

We propose two HTE-based feature selection methods for CMAB: Heterogeneous Incremental Effect (HIE) and Heterogeneous Distribution Divergence (HDD). For simplicity, we consider a binary reward $Y$, where $P_i(1)$ is the global probability of reward for arm $i$, and $P_{b_i}(1)$ is the probability of reward for arm $i$ within bin $b$ of a feature $x$. Continuous features are discretized into $m$ bins (e.g., equal sample size), each bin $b$ containing $N_b$ samples. Categorical features use their inherent categories.

\subsection{Heterogeneous Incremental Effect (HIE) Score}
A feature $X$ is important if the local best arm in some feature bins differs from the global best arm. The HIE score captures the incremental gain in reward from selecting the local best arm in feature bins compared to selecting the global best arm.
The HIE score for feature $x$ with $m$ bins is:
\begin{align*}
FI_{HIE}(x|m) &= \sum_{b=1}^{m} \frac{N_b}{N} \left[P_{w_{b}}(1) - P_{w^{*}}(1)\right] \\
&= \sum_{b=1}^{m} \frac{N_b}{N} \left[\max_{i \in \{1,...,k\}}P_{b_i}(1) - \max_{i \in \{1,...,k\}}P_{i}(1)\right]
\end{align*}
where $w_{b} := \arg\max_{i} P_{b_i}(1)$ is the local best arm in bin $b$, and $w^{*} := \arg\max_{i} P_{i}(1)$ is the global best arm.

\begin{proposition}
\label{prop:HIE-nonnegative}
The HIE score is non-negative: $FI_{HIE}(x|m) \geq 0$.
\end{proposition}
Proposition \ref{prop:HIE-nonnegative} (proof in Appendix \ref{proof:HIE-nonnegative}) formally establishes the non-negative nature of the HIE score, ensuring it can be interpreted as a magnitude of incremental effect.

\begin{proposition}
\label{prop:HIE-bias}
The expected value of the HIE score increases as the number of bins increases:
\[
\mathbb{E}[FI_{HIE}(x \mid m + i)] \;\ge\; \mathbb{E}[FI_{HIE}(x \mid m)]
\quad (i>0),
\]
provided that the additional bins are created by splitting the existing $m$ bins.
\end{proposition}
As shown in Proposition \ref{prop:HIE-bias} (proof in Appendix \ref{proof:HIE-bias}), the unnormalized HIE score tends to increase with finer binning. This observation motivates our introduction of a normalized score to mitigate this potential bias when comparing features binned differently or to assess significance.

\subsubsection{Normalized HIE Score}
To reduce bias from varying bin counts and establish a baseline under the null hypothesis (feature and reward are independent), we introduce a normalized HIE score. The normalization offsets the expected mean of HIE under the null, estimated via bootstrap sampling (randomly splitting data into $m$ bins with original sizes for $S$ trials).
The normalized HIE score is:
\begin{align*}
FI_{NHIE}(x \mid m)
&= \sum_{b=1}^{m} \frac{N_b}{N} \max_{i \in \{1,\dots,k\}} P_{b_i}(1)
\;-\;
\frac{1}{S} \sum_{s=1}^{S} \sum_{b=1}^{m} \frac{N_b}{N} \max_{i \in \{1,\dots,k\}} P_{b_{is}}(1),
\end{align*}
where $P_{b_{is}}(1)$ is the positive label probability for arm $i$ in bin $b$ of the $s$-th bootstrap trial.

\subsubsection{Feature Importance Statistical Significance}
Statistical significance is evaluated using a bootstrap-based p-value for the normalized HIE score:
\begin{align*}
p_{NHIE} = \frac{1}{S} \sum_{s=1}^{S} \mathbb{I} \left( \sum_{b=1}^{m} \frac{N_b}{N} \max_{i} P_{b_i}(1) \leq \sum_{b=1}^{m} \frac{N_b}{N} \max_{i} P_{b_{is}}(1) \right).
\end{align*}

\subsubsection{Algorithm: Computing Normalized HIE Score and p-value}
\label{sec:algo_nhie}
Algorithm~\ref{algo:NHIE} details this procedure.

\begin{algorithm}
\caption{Normalized HIE Score \& P-value}
\label{algo:NHIE}
\begin{algorithmic}[1]
\REQUIRE Data $\mathcal{D}$, bins $m$, bootstrap samples $S$
\ENSURE $FI_{NHIE}$, $p_{NHIE}$

\STATE Compute $\Psi_{\text{obs}} \gets \sum_{b=1}^m \frac{N_b}{N} \max_i P_{b_i}(1)$ \COMMENT{Observed sum of max reward probabilities}
\STATE Initialize $\{\Psi_s\}_{s=1}^S \gets \emptyset$ \COMMENT{Bootstrap scores storage}
\FOR{$s \in 1:S$}
    \STATE Shuffle $\mathcal{D}$ preserving bin sizes
    \STATE $\Psi_s \gets \sum_{b=1}^m \frac{N_b}{N} \max_i \hat{P}_{b_i}^{(s)}(1)$ \COMMENT{Bootstrap sample score}
    \STATE Store $\Psi_s$ in $\{\Psi_s\}$
\ENDFOR
\STATE $\mu_{\text{null}} \gets \frac{1}{S}\sum_{s=1}^S \Psi_s$ \COMMENT{Null distribution mean}
\STATE $FI_{NHIE} \gets \Psi_{\text{obs}} - \mu_{\text{null}}$ \COMMENT{Normalized score}
\STATE $p_{NHIE} \gets \frac{1}{S}\sum_{s=1}^S \mathbb{I}(\Psi_{\text{obs}} \leq \Psi_s)$ \COMMENT{Right-tail p-value}
\RETURN $(FI_{NHIE}, p_{NHIE})$
\end{algorithmic}
\end{algorithm}

Note: The original Algorithm 1 calculates $FI_{HIE}$ then normalizes it. The revised Algorithm 1 above calculates the observed term of HIE (first sum), normalizes this term by its bootstrap mean, and then the raw HIE can be recovered by subtracting $\max_i P_i(1)$ if needed. The p-value is based on the observed term. This is consistent with the formula for $FI_{NHIE}$ where $\max_i P_i(1)$ is effectively removed from both terms before bootstrap.

\subsection{Heterogeneous Distribution Divergence (HDD) Score}
The HDD score quantifies heterogeneity in reward distributions across bins using KL divergence.
\begin{align*}
FI_{HDD}(x|m) = \sum_{b=1}^{m} \frac{N_b}{N} D_b(P_{b_1}, ..., P_{b_k}) - D(P_{1}, ..., P_{k}),
\end{align*}
where $D_b$ represents the average pairwise Kullback-Leibler (KL) divergence between the reward distributions of all arm pairs within bin $b$, and $D$ is the corresponding global divergence across the entire dataset. Higher $D_b$ values indicate greater heterogeneity in arm outcomes within the bin. While theoretically non-negative (as true HTE features should increase local divergence), small negative values may occur due to finite-sample estimation errors. The score generally increases with finer binning ($m$) when true HTE exists, though excessive binning can introduce noise. (See Appendix~\ref{sec:appendix_hdd} for full mathematical formulation.)

\subsubsection{Normalized HDD Score}
To reduce bias from varying bin counts:
\begin{align*}
FI_{NHDD}(x|m) = \sum_{b=1}^{m} \frac{N_b}{N} D_b(P_{b_1}, ..., P_{b_k})
 - \frac{1}{S} \sum_{s=1}^{S} \sum_{b=1}^{m} \frac{N_b}{N} D_b(P_{b_{1s}}, ..., P_{b_{ks}}).
\end{align*}

\subsubsection{Statistical Significance of the HDD Score}
The p-value is computed using bootstrap samples:
\begin{align*}
p_{NHDD} = \frac{1}{S} \sum_{s=1}^{S} \mathbb{I}\left[\sum_{b=1}^{m} \frac{N_b}{N} D_b(P_{b_1}, ..., P_{b_k}) \leq \sum_{b=1}^{m} \frac{N_b}{N} D_b(P_{b_{1s}}, ..., P_{b_{ks}})\right].
\end{align*}

\subsubsection{Algorithm: Computing Normalized HDD Score and p-value}
\label{sec:algo_nhdd}
Algorithm~\ref{algo:NHDD} outlines this procedure.
\begin{algorithm}
\caption{Normalized HDD Score \& P-value}
\label{algo:NHDD}
\begin{algorithmic}[1]
\REQUIRE Data $\mathcal{D}$, bins $m$, bootstrap samples $S$
\ENSURE $FI_{NHDD}$, $p_{NHDD}$

\STATE Compute $\Phi_{\text{obs}} \gets \sum_{b=1}^m \frac{N_b}{N} D_b(\{P_{b_i}\}_{i=1}^k)$ \COMMENT{Observed divergence sum}
\STATE Initialize $\{\Phi_s\}_{s=1}^S \gets \emptyset$ \COMMENT{Bootstrap scores storage}
\FOR{$s \in 1:S$}
    \STATE Shuffle $\mathcal{D}$ preserving bin sizes
    \STATE $\Phi_s \gets \sum_{b=1}^m \frac{N_b}{N} D_b(\{\hat{P}_{b_i}^{(s)}\}_{i=1}^k)$ \COMMENT{Bootstrap sample divergence}
    \STATE Store $\Phi_s$ in $\{\Phi_s\}$
\ENDFOR
\STATE $\mu_{\text{null}} \gets \frac{1}{S}\sum_{s=1}^S \Phi_s$ \COMMENT{Null distribution mean}
\STATE $FI_{NHDD} \gets \Phi_{\text{obs}} - \mu_{\text{null}}$ \COMMENT{Normalized divergence score}
\STATE $p_{NHDD} \gets \frac{1}{S}\sum_{s=1}^S \mathbb{I}(\Phi_{\text{obs}} \leq \Phi_s)$ \COMMENT{Right-tail p-value}
\RETURN $(FI_{NHDD}, p_{NHDD})$
\end{algorithmic}
\end{algorithm}

\section{Evaluation}
The empirical performance of these two feature importance scores is compared in this section. Normalized scores are used, with $S=100$ bootstrap trials.

\subsection{Evaluation with Synthetic Data}
\label{sec:synthetic_evaluation}

\begin{figure*}[h]
\centering
\includegraphics[width=1.0 \textwidth]{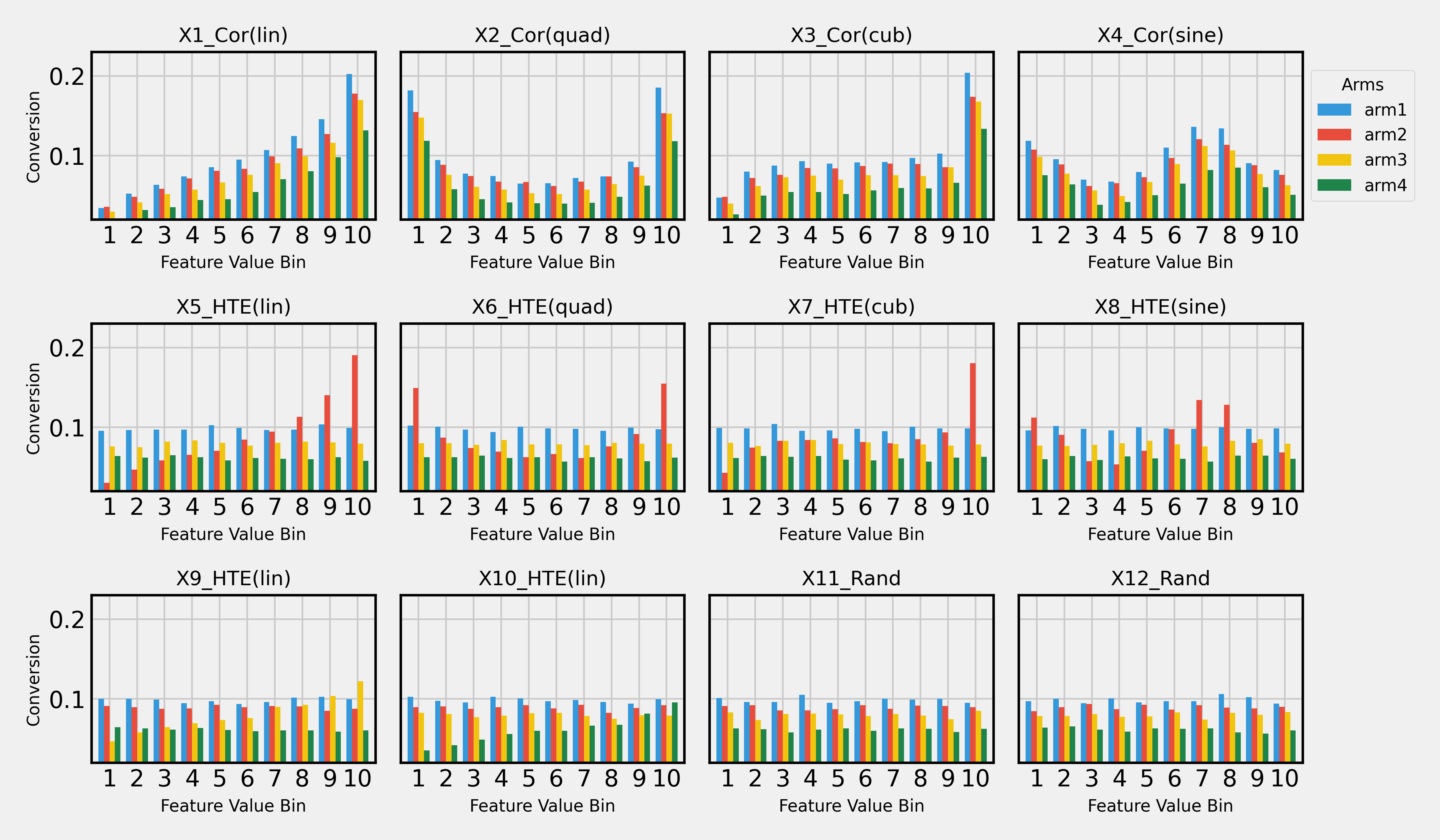} 
\caption{Illustration of feature patterns in synthetic data (12 features, 4 arms, N=100k, visualized with 10 bins). HTE features (e.g., X5-X9) alter relative arm performance, unlike purely correlational (X1-X4) or random (X11-X12) features.}
\label{fig:feature_pattern}
\end{figure*}

\begin{figure*}[htbp] 
    \centering 
    
    \begin{subfigure}[b]{\linewidth} 
        \centering 
        \includegraphics[width=0.9\textwidth]{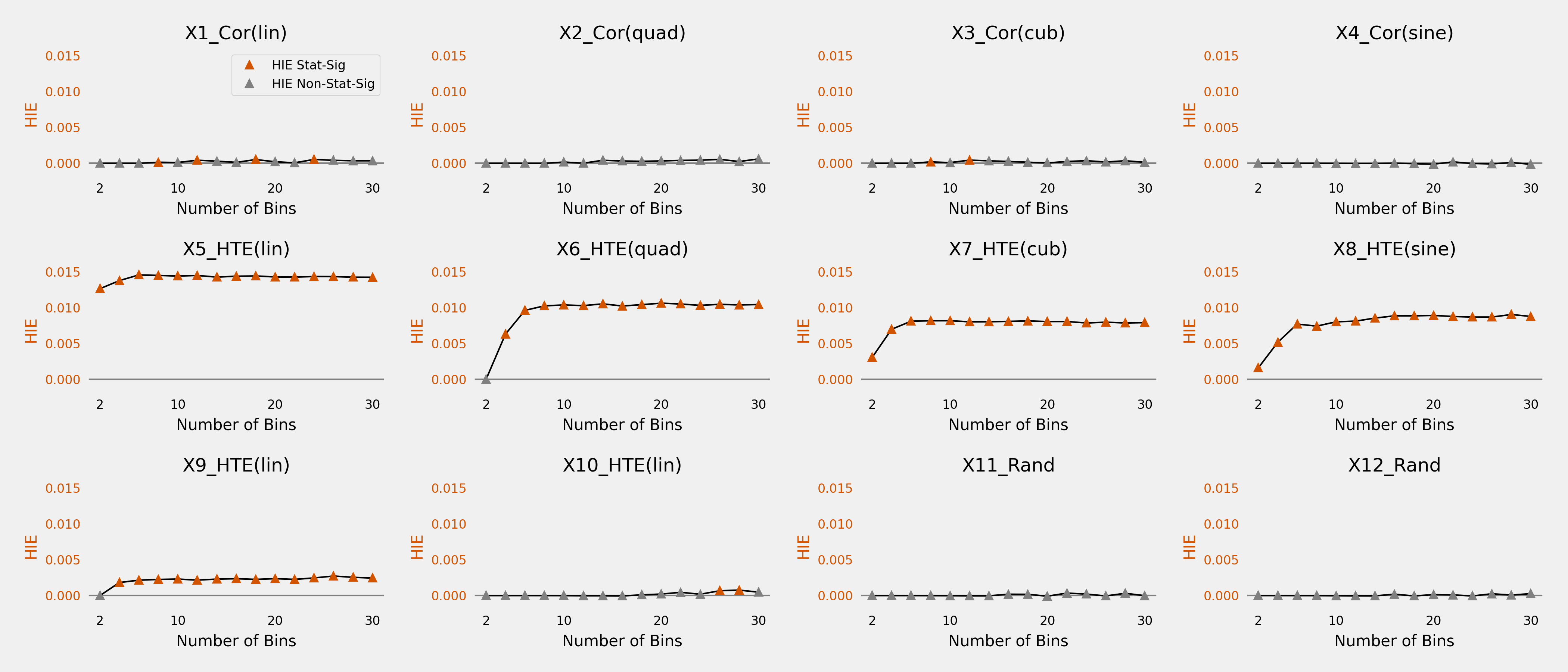} 
        \caption{HIE Score vs. Bin Count}
        \label{fig:bin_selection_hie}
    \end{subfigure}
    

    \begin{subfigure}[b]{\linewidth} 
        \centering 
        \includegraphics[width=0.9\textwidth]{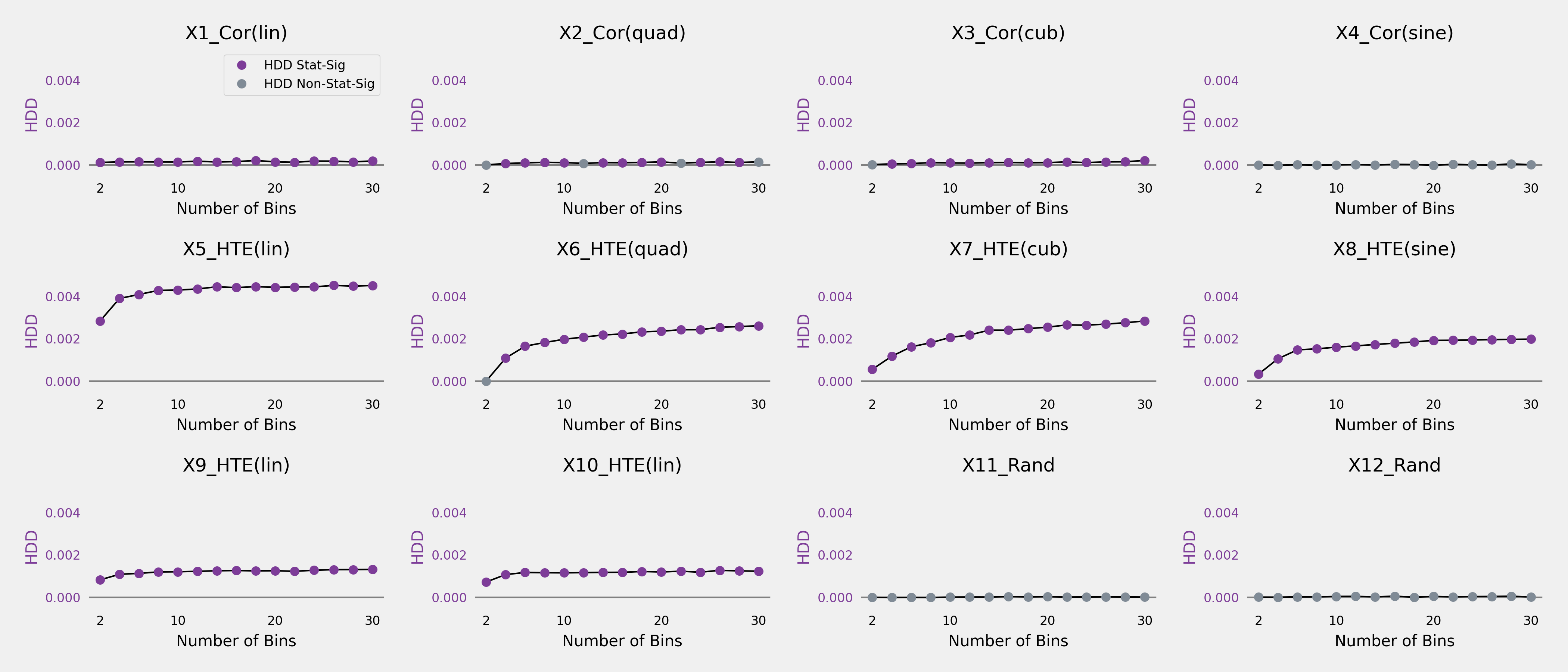} 
        \caption{HDD Score vs. Bin Count}
        \label{fig:bin_selection_hdd}
    \end{subfigure}
    
    \caption{Sensitivity of normalized HIE and HDD scores to the number of bins ($m$) on synthetic data (N=100,000). Statistical significance of scores is color-coded. (a) HIE scores. (b) HDD scores.}
    \label{fig:bin_selection} 
\end{figure*}

\begin{figure*}[h] 
\includegraphics[width=\linewidth]{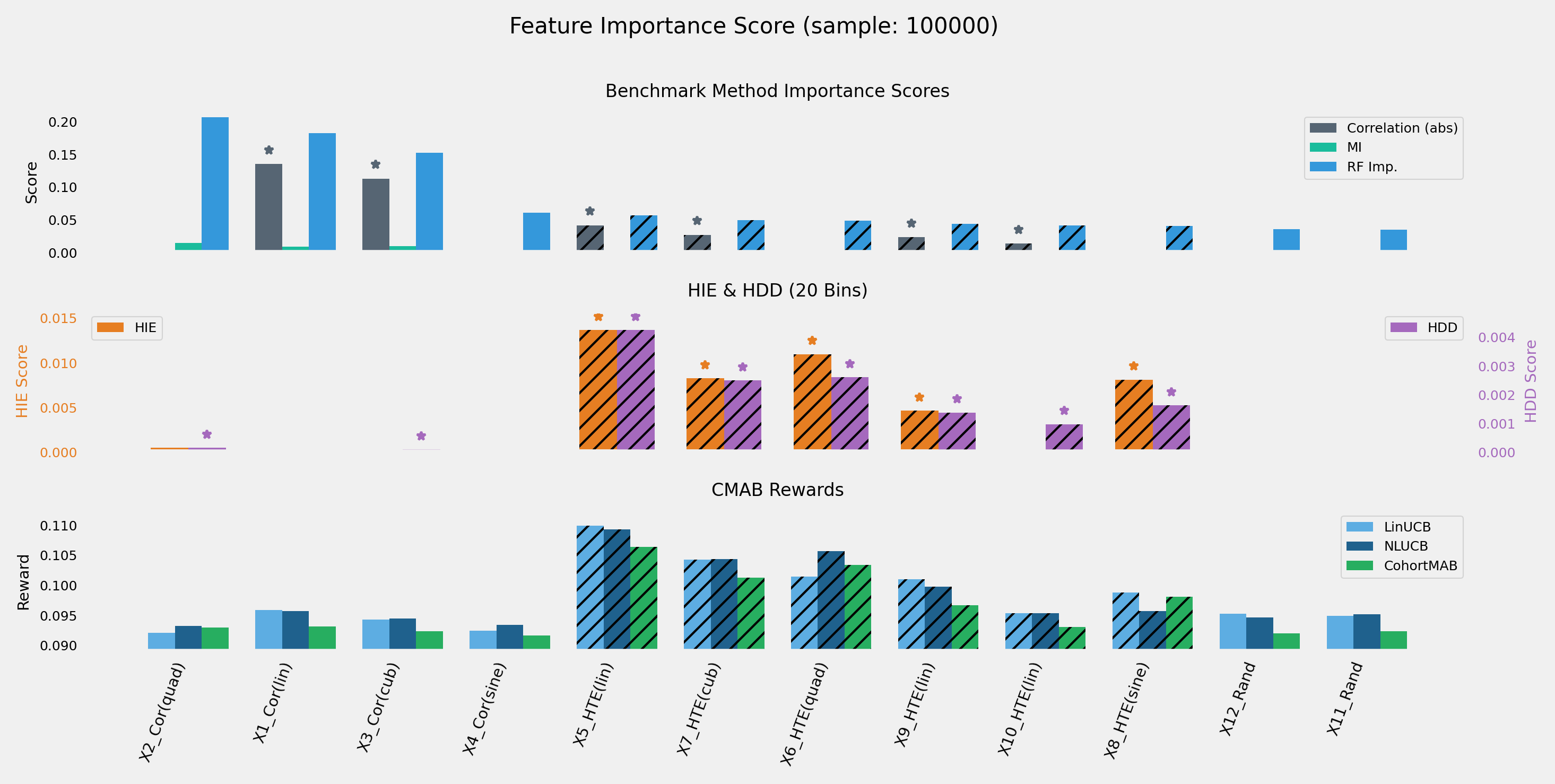}
\caption{Comparison of feature importance scores (HIE, HDD, Pearson Correlation) and CMAB rewards (LinUCB, NonLinearUCB, CohortMAB) for synthetic data at N=100,000 (20 bins for HIE/HDD). Stars indicate statistical significance (p<0.01) for applicable methods (Correlation, HIE, HDD). True HTE features (X5-X9, with hatch pattern) are consistently highly ranked by HIE/HDD and yield high CMAB rewards.}
\label{fig:synthetic_comparison_scores_rewards}
\end{figure*}

\begin{table*}[htbp] 
\centering
\caption{Aggregated Evaluation Summary Across 10 Trials (Mean \textpm{} Std). Best scores per metric and sample size are in bold. HIE/HDD use $n_{bins}=20, p<0.1$. P@6/R@6 denote Precision/Recall at 6 features.}
\label{tab:aggregated_results}
\setlength{\tabcolsep}{4pt} 
\begin{tabular}{@{}ll S[table-format=1.3,table-figures-uncertainty=1]
                           S[table-format=1.3,table-figures-uncertainty=1]
                           S[table-format=1.3,table-figures-uncertainty=1]
                           S[table-format=1.3,table-figures-uncertainty=1]
                           S[table-format=1.3,table-figures-uncertainty=1]
                           S[table-format=1.3,table-figures-uncertainty=1]@{}}
\toprule
\multirow{2}{*}{Samples} & \multirow{2}{*}{Method} & \multicolumn{2}{c}{AUC-PR} & \multicolumn{2}{c}{Precision@6} & \multicolumn{2}{c}{Recall@6} \\
\cmidrule(lr){3-4} \cmidrule(lr){5-6} \cmidrule(lr){7-8}
 & & {Mean} & {Std} & {Mean} & {Std} & {Mean} & {Std} \\
\midrule
\multirow{8}{*}{1,000}   & HDD                     & \bfseries 0.813 & 0.112 & 0.433 & 0.179 & 0.433 & 0.179 \\
                         & HIE                     & 0.782 & 0.146 & 0.417 & 0.212 & 0.417 & 0.212 \\
                         & Pearson Correlation (abs) & 0.441 & 0.047 & 0.517 & 0.146 & 0.517 & 0.146 \\
                         & Mutual Information      & 0.374 & 0.041 & 0.367 & 0.105 & 0.367 & 0.105 \\
                         & Random Forest Imp.      & 0.362 & 0.019 & 0.317 & 0.095 & 0.317 & 0.095 \\
                         & CohortMAB Reward        & 0.623 & 0.181 & \bfseries 0.533 & 0.189 & \bfseries 0.533 & 0.189 \\
                         & LinUCB Reward           & 0.534 & 0.145 & 0.483 & 0.146 & 0.483 & 0.146 \\
                         & NonLinear LinUCB Reward & 0.632 & 0.159 & 0.583 & 0.196 & 0.583 & 0.196 \\
\addlinespace
\multirow{8}{*}{10,000}  & HDD                     & \bfseries 0.997 & 0.008 & \bfseries 0.983 & 0.053 & \bfseries 0.983 & 0.053 \\
                         & HIE                     & 0.953 & 0.044 & 0.867 & 0.105 & 0.867 & 0.105 \\
                         & Pearson Correlation (abs) & 0.477 & 0.036 & 0.650 & 0.053 & 0.650 & 0.053 \\
                         & Mutual Information      & 0.350 & 0.023 & 0.283 & 0.081 & 0.283 & 0.081 \\
                         & Random Forest Imp.      & 0.380 & 0.012 & 0.333 & 0.000 & 0.333 & 0.000 \\
                         & CohortMAB Reward        & 0.872 & 0.067 & 0.767 & 0.117 & 0.767 & 0.117 \\
                         & LinUCB Reward           & 0.921 & 0.056 & 0.833 & 0.111 & 0.833 & 0.111 \\
                         & NonLinear LinUCB Reward & 0.868 & 0.083 & 0.750 & 0.118 & 0.750 & 0.118 \\
\addlinespace
\multirow{8}{*}{100,000} & HDD                     & \bfseries 1.000 & 0.000 & \bfseries 1.000 & 0.000 & \bfseries 1.000 & 0.000 \\
                         & HIE                     & 0.954 & 0.031 & 0.867 & 0.070 & 0.867 & 0.070 \\
                         & Pearson Correlation (abs) & 0.473 & 0.028 & 0.667 & 0.000 & 0.667 & 0.000 \\
                         & Mutual Information      & 0.355 & 0.027 & 0.300 & 0.070 & 0.300 & 0.070 \\
                         & Random Forest Imp.      & 0.386 & 0.000 & 0.333 & 0.000 & 0.333 & 0.000 \\
                         & CohortMAB Reward        & 0.942 & 0.035 & 0.850 & 0.053 & 0.850 & 0.053 \\
                         & LinUCB Reward           & 0.971 & 0.026 & 0.900 & 0.086 & 0.900 & 0.086 \\
                         & NonLinear LinUCB Reward & 0.961 & 0.032 & 0.867 & 0.105 & 0.867 & 0.105 \\
\bottomrule
\end{tabular}
\end{table*}

We evaluate our proposed feature selection methods using synthetic data generated via the \textit{CausalML} Python package \cite{chen2020causalml, zhao2022feature}. The dataset comprises $100,000$ samples, $4$ arms, and $12$ features meticulously designed to exhibit diverse characteristics: $4$ features purely correlated with the outcome without inducing HTE (linear, quadratic, cubic, sine correlations); $6$ HTE features with varying patterns (linear, quadratic, cubic, sine, and two weaker linear HTEs, one of which only shifts distributions without altering the local best arm); and $2$ random irrelevant features. Figure~\ref{fig:feature_pattern} illustrates these feature patterns, where continuous features are discretized into 10 decile-based bins for visualization, highlighting the global best arm (arm 1) and instances of local best arms indicating heterogeneity.

We compare HIE and HDD against several benchmarks: traditional methods (Pearson Correlation, Mutual Information, Random Forest Importance) and model-embedded MAB approaches where feature importance is derived from running MABs (LinUCB, NonLinearUCB, CohortMAB) with individual features as context. The MAB rewards are obtained using replay evaluation \cite{li2011unbiased}. 

Table~\ref{tab:aggregated_results} summarizes the performance across 10 trials for key sample sizes ($N \in \{1,000, 10,000, 100,000\}$), using AUC-PR, Precision@6, and Recall@6 as evaluation metrics (assuming 6 true HTE features). **Results for intermediate sample sizes ($N \in \{5,000, 50,000\}$) showed consistent trends with the reported values and are included in Appendix~\ref{app:full_synthetic_results} for completeness.

Our proposed methods, HIE and HDD, demonstrate superior performance in identifying true HTE features. As shown in Table~\ref{tab:aggregated_results}, HDD consistently achieves the highest AUC-PR, reaching a perfect score of $1.000 \pm 0.000$ for $N \ge 50,000$. HIE also performs strongly, with an AUC-PR of $0.954 \pm 0.031$ at $N=100,000$. Both significantly outperform traditional methods like Pearson Correlation (AUC-PR $0.473 \pm 0.028$ at $N=100,000$) and Random Forest Importance (AUC-PR $0.386 \pm 0.000$ at $N=100,000$). Notably, even MAB-derived feature importances (e.g., LinUCB Reward achieving AUC-PR $0.971 \pm 0.026$ at $N=100,000$) are slightly edged out by HDD. This underscores the efficacy of directly targeting HTE signals. Both HIE and HDD effectively capture non-linear HTE patterns where methods like LinUCB, being model-based, falter if the linearity assumption is violated, as evidenced by its lower reward on features with quadratic or cubic HTE. For instance, Figure~\ref{fig:synthetic_comparison_scores_rewards} (for $N=100,000$) visually confirms the alignment of high HIE/HDD scores with features yielding high CMAB rewards, particularly for true HTE features ($X5$ to $X9$). Conversely, purely correlational features ($X1$ to $X4$) receive low HIE/HDD scores despite their correlation with the outcome, highlighting the advantage of our HTE-focused methods over standard correlation metrics.

The choice of bin count, $m$, is a crucial hyperparameter. Figure~\ref{fig:bin_selection} illustrates the sensitivity of normalized HIE and HDD scores to $m$, based on experiments with $N=100,000$. A minimum number of bins (e.g., $m \ge 6$) is generally required to effectively detect HTE patterns, especially non-linear ones. Performance tends to saturate or show diminishing returns with very fine-grained binning (e.g., $m > 25-30$), where bins may contain too few samples, increasing score variance. For the features in our synthetic dataset, a range of $m \in [15, 25]$ appears robust for both HIE and HDD.

Comparing HIE and HDD, HDD tends to be more sensitive to any distributional shift caused by a feature, even if it doesn't change the local best arm. For example, the weak HTE feature $X10\_HTE(lin)$, which shifts distributions but not the winning arm in our 10-bin visualization, shows statistical significance for HDD but not consistently for HIE (Table~\ref{tab:aggregated_results}, Figure~\ref{fig:synthetic_comparison_scores_rewards}). This suggests HDD might be more suitable for exploratory analysis aiming to find any feature influence, while HIE is more directly tied to features impacting optimal arm selection and immediate reward uplift. Both methods correctly assign low importance to random and purely correlational features. Direct comparison to feature importances from complex HTE models like causal forests was beyond the scope of this work, which focuses on computationally lean filter methods.

\subsection{Online Experiments}
\label{sec:online_experiments}

We validated our methods within a large-scale recommender system, specifically for the Home page thumbnail personalization feature. This system aims to enhance user engagement by displaying the most relevant thumbnail for each experience (content item) to individual users, optimizing for user conversions. For each content, creators can activate multiple thumbnails (typically 2-5), and the platform's underlying bandit algorithm dynamically allocates impressions to optimize the reward, balancing exploration of different thumbnails with exploitation of high-performing ones for different user segments defined by the contextual feature.
Our goal was to select impactful contextual features at a system level to enhance these experience-specific CMABs, where each experience is a separate CMAB instance and its various thumbnails are the arms. 

\begin{figure*}[h] 
\centering
\includegraphics[width=\linewidth]{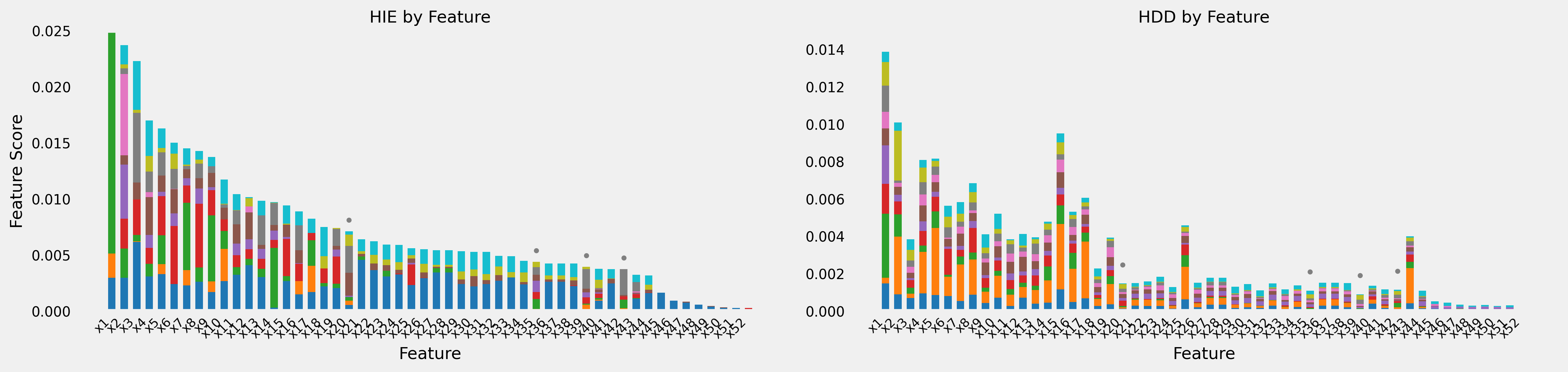}
\caption{Feature scores based on online non-contextual MAB data: Features are ordered by the HIE score. Each color represents a content item, and four random benchmark features are indicated with grey dots.}
\label{fig:real_data_feature_selection}
\end{figure*}

The validation involved several steps:
First, a platform-wide non-contextual MAB experiment was conducted to gather baseline user-impression data across diverse content items and thumbnails. This provided the training data for our feature selection methods.
Second, leveraging this data, HIE and HDD were applied offline to a broad pool of candidate features (Figure~\ref{fig:real_data_feature_selection}), resulting in a refined short-list of high-importance features. This offline pre-selection was critical, as exhaustively testing all candidate features in a live environment serving tens of millions of users daily would be prohibitively expensive and complex. Figure~\ref{fig:real_data_feature_selection} shows cumulative HIE scores for 52 features (48 real, 4 random benchmark) from 97 sample content items. Feature $X1$ is synthetic (concatenation of $X2, X3$). Random features rank low, suggesting features with comparable scores are unlikely to be informative.

Third, an online A/B experiment was deployed to evaluate the CMAB performance using these selected features ($X1, X2, X3$). The control group utilized a non-contextual MAB, while treatment groups implemented CMABs (cohort-based Thompson Sampling) using one of these top features. Each user received personalized homepage recommendations, with thumbnail selections for participating content items determined by their assigned experimental condition.

The online experiment results are summarized in Table~\ref{tab:online_experiment_summary}. Feature $X1$, which had a high offline HIE score, demonstrated the most statistically significant wins (57) and the highest cumulative reward gain (0.774) out of 330 content items when used as context in a CMAB, compared to the non-contextual MAB baseline. This aligns with its ability to predict features leading to significant CMAB treatment effects ($p<0.01$), as indicated by an ROC AUC of 0.81 when ranking features by HIE p-values (using scores for ties). Features $X2$ and $X3$ also showed positive impact, with $X3$ having a slightly higher HDD-based ROC AUC (0.74) despite fewer significant wins. While this online setup doesn't directly compare different \textit{feature selection methods} due to operational costs, the strong correlation between offline HIE/HDD metrics (particularly for X1) and superior online CMAB performance provides compelling evidence for our methods' effectiveness in identifying impactful features for real-world deployment.

\begin{table}[htbp]
\centering
\caption{Online Experiment Summary: Performance of CMABs using selected features (X1, X2, X3) against a non-contextual MAB baseline (330 content items), and ROC AUC for identifying features leading to significant CMAB treatment effects (based on HIE/HDD p-values, scores for ties).}
\label{tab:online_experiment_summary}
\begin{tabular}{@{}lcccc@{}}
\toprule
CMAB  & Significant Wins  & Reward  & HIE & HDD \\
Feature & ($p<0.01$) & Gain & (ROC AUC) & (ROC AUC) \\
\midrule
X1 & 57 & 0.774 & 0.81 & 0.72 \\
X2 & 46 & 0.608 & 0.78 & 0.72 \\
X3 & 43 & 0.484 & 0.61 & 0.74 \\
\bottomrule
\end{tabular}
\end{table}

\subsection{Computational Considerations and Practical Guidance}
\label{sec:practical_considerations}
\textbf{Computational Complexity}: The HIE and HDD scores are computationally efficient. For $N$ samples, $k$ arms, $m$ bins, and $S$ bootstrap trials, the overall complexity is roughly $O(N + S \cdot m \cdot k)$. This makes them substantially faster than methods requiring retraining complex models for feature importance.

\textbf{Scalability}: Our methods have been successfully applied in a system with tens of millions of daily user impressions (implying large N) and evaluating features for CMABs across thousands of content items. The model-free nature and efficient per-feature scoring contribute to their scalability for screening a large candidate feature pool, as demonstrated in our online deployment.

\textbf{Redundant Features}: As filter methods, HIE and HDD may assign high scores to multiple correlated features capturing similar underlying HTE. In practice, this can be addressed by standard post-processing, such as selecting the feature with the highest score from a highly correlated cluster. Future work could explore integrating redundancy penalties.

\textbf{Bin Count ('m')}: The choice of bin count is a key hyperparameter. Figures \ref{fig:bin_selection_hie} and \ref{fig:bin_selection_hdd} illustrate score sensitivity to 'm'. Practitioners might explore a small range of 'm', use domain knowledge for discretization, and the choice might also depend on the expected granularity of HTE. 

\textbf{Combining HIE/HDD Scores}: HIE and HDD offer complementary perspectives. HIE measures incremental reward gain from heterogeneous arm selection, relevant for immediate impact. HDD is sensitive to any distributional shift, potentially identifying subtle HTE. Practitioners may prioritize features high on both, use HIE for strong HTE confirmation, and HDD for exploratory discovery. 

\section{Conclusion}
This paper introduced HIE and HDD, two novel, model-free filter methods designed to identify features indicative of heterogeneous treatment effects in CMABs, a crucial step for effective personalization in large-scale systems. Unlike traditional correlation-based approaches, our methods directly quantify how features contribute to variations in optimal arm selection and reward distributions. Synthetic data experiments demonstrated their ability to capture diverse HTE patterns and their complementary strengths. The successful deployment and validation within a large-scale commercial recommender system, where features selected by HIE and HDD led to significant CMAB performance improvements, underscore their practical utility and computational efficiency. These methods provide a robust and interpretable approach to feature selection, mitigating risks of model mis-specification inherent in model-based techniques.

Future research directions include developing systematic hyperparameter selection strategies for binning, conducting more extensive comparisons with a wider array of feature selection baselines across diverse datasets, and exploring extensions to explicitly model feature interactions. Further investigation into integrating formal causal inference techniques to adjust for potential unobserved confounding, and adapting these methods for settings with extremely high-dimensional sparse features or dynamic, non-stationary environments, would also be valuable.

\section*{References}
\small
\bibliographystyle{ACM-Reference-Format} 
\bibliography{reference} 

@inproceedings{li2010contextual,
  title={A contextual-bandit approach to personalized news article recommendation},
  author={Li, Lihong and Chu, Wei and Langford, John and Schapire, Robert E},
  booktitle={Proceedings of the 19th International Conference on World Wide Web},
  pages={661--670},
  year={2010}
}

@inproceedings{zhao2022feature,
  title={Feature selection methods for uplift modeling and heterogeneous treatment effect},
  author={Zhao, Zhenyu and Zhang, Yumin and Harinen, Totte and Yung, Mike},
  booktitle={IFIP International Conference on Artificial Intelligence Applications and Innovations},
  pages={217--230},
  year={2022},
  organization={Springer}
}

@inproceedings{bouneffouf2024tutorial,
  title={A Tutorial on Multi-Armed Bandit Applications for Large Language Models},
  author={Bouneffouf, Djallel and F{\'e}raud, Rapha{\"e}l},
  booktitle={Proceedings of the 30th ACM SIGKDD Conference on Knowledge Discovery and Data Mining},
  pages={6412--6413},
  year={2024}
}

@article{slivkins2019introduction,
  title={Introduction to multi-armed bandits},
  author={Slivkins, Aleksandrs and others},
  journal={Foundations and Trends{\textregistered} in Machine Learning},
  volume={12},
  number={1-2},
  pages={1--286},
  year={2019},
  publisher={Now Publishers, Inc.}
}

@inproceedings{heyden2024budgeted,
  title={Budgeted Multi-Armed Bandits with Asymmetric Confidence Intervals},
  author={Heyden, Marco and Arzamasov, Vadim and Fouch{\'e}, Edouard and B{\"o}hm, Klemens},
  booktitle={Proceedings of the 30th ACM SIGKDD Conference on Knowledge Discovery and Data Mining},
  pages={1073--1084},
  year={2024}
}

@article{gangan9survey,
  title={Survey of multiarmed bandit algorithms applied to recommendation systems},
  author={Gangan, Elena and Kudus, Milos and Ilyushin, Eugene},
  journal={International Journal of Open Information Technologies},
  volume={9},
  number={4},
  pages={2021}
}

@misc{kawale2018multiarmed,
  title={A MultiArmed Bandit Framework For Recommendations at Netflix},
  author={Kawale, Jaya and Chow, Elliot},
  year={2018}
}

@inproceedings{li2011unbiased,
  title={Unbiased offline evaluation of contextual-bandit-based news article recommendation algorithms},
  author={Li, Lihong and Chu, Wei and Langford, John and Wang, Xuanhui},
  booktitle={Proceedings of the fourth ACM international conference on Web search and data mining},
  pages={297--306},
  year={2011}
}

@inproceedings{mcinerney2018explore,
  title={Explore, exploit, and explain: personalizing explainable recommendations with bandits},
  author={McInerney, James and Lacker, Benjamin and Hansen, Samantha and Higley, Karl and Bouchard, Hugues and Gruson, Alois and Mehrotra, Rishabh},
  booktitle={Proceedings of the 12th ACM conference on recommender systems},
  pages={31--39},
  year={2018}
}

@inproceedings{yi2023online,
  title={Online Matching: A Real-time Bandit System for Large-scale Recommendations},
  author={Yi, Xinyang and Wang, Shao-Chuan and He, Ruining and Chandrasekaran, Hariharan and Wu, Charles and Heldt, Lukasz and Hong, Lichan and Chen, Minmin and Chi, Ed H},
  booktitle={Proceedings of the 17th ACM Conference on Recommender Systems},
  pages={403--414},
  year={2023}
}

@article{wager2018estimation,
  title={Estimation and inference of heterogeneous treatment effects using random forests},
  author={Wager, Stefan and Athey, Susan},
  journal={Journal of the American Statistical Association},
  volume={113},
  number={523},
  pages={1228--1242},
  year={2018},
  publisher={Taylor \& Francis}
}

@article{foster2019model,
  title={Model selection for contextual bandits},
  author={Foster, Dylan J and Krishnamurthy, Akshay and Luo, Haipeng},
  journal={Advances in Neural Information Processing Systems},
  volume={32},
  year={2019}
}

@inproceedings{zhao2019maximum,
  title={Maximum relevance and minimum redundancy feature selection methods for a marketing machine learning platform},
  author={Zhao, Zhenyu and Anand, Radhika and Wang, Mallory},
  booktitle={2019 IEEE international conference on data science and advanced analytics (DSAA)},
  pages={442--452},
  year={2019},
  organization={IEEE}
}

@article{agrawal2021metaheuristic,
  title={Metaheuristic algorithms on feature selection: A survey of one decade of research (2009-2019)},
  author={Agrawal, Prachi and Abutarboush, Hattan F and Ganesh, Talari and Mohamed, Ali Wagdy},
  journal={Ieee Access},
  volume={9},
  pages={26766--26791},
  year={2021},
  publisher={IEEE}
}

@article{theng2024feature,
  title={Feature selection techniques for machine learning: a survey of more than two decades of research},
  author={Theng, Dipti and Bhoyar, Kishor K},
  journal={Knowledge and Information Systems},
  volume={66},
  number={3},
  pages={1575--1637},
  year={2024},
  publisher={Springer}
}

@article{chen2020causalml,
  title={Causalml: Python package for causal machine learning},
  author={Chen, Huigang and Harinen, Totte and Lee, Jeong-Yoon and Yung, Mike and Zhao, Zhenyu},
  journal={arXiv preprint arXiv:2002.11631},
  year={2020}
}

@article{bolon2013review,
  title={A review of feature selection methods on synthetic data},
  author={Bol{\'o}n-Canedo, Ver{\'o}nica and S{\'a}nchez-Maro{\~n}o, Noelia and Alonso-Betanzos, Amparo},
  journal={Knowledge and information systems},
  volume={34},
  number={3},
  pages={483--519},
  year={2013},
  publisher={Springer}
}

@article{chandrashekar2014survey,
  title={A survey on feature selection methods},
  author={Chandrashekar, Girish and Sahin, Ferat},
  journal={Computers \& Electrical Engineering},
  volume={40},
  number={1},
  pages={16--28},
  year={2014},
  publisher={Elsevier}
}

@inproceedings{lin2022adafs,
  title={AdaFS: Adaptive feature selection in deep recommender system},
  author={Lin, Weilin and Zhao, Xiangyu and Wang, Yejing and Xu, Tong and Wu, Xian},
  booktitle={Proceedings of the 28th ACM SIGKDD Conference on Knowledge Discovery and Data Mining},
  pages={3309--3317},
  year={2022}
}

@article{lai1985asymptotically,
  title={Asymptotically efficient adaptive allocation rules},
  author={Lai, Tze Leung and Robbins, Herbert},
  journal={Advances in applied mathematics},
  volume={6},
  number={1},
  pages={4--22},
  year={1985},
  publisher={Academic Press}
}

@inproceedings{tang2014ensemble,
  title={Ensemble contextual bandits for personalized recommendation},
  author={Tang, Lixin and Wang, Shuaiqiang and Aggarwal, Charu C},
  booktitle={Proceedings of the 8th ACM Conference on Recommender systems (RecSys '14)},
  pages={109--116},
  year={2014},
  organization={ACM}
}

@article{wang2023policy,
  title={Policy-Guided Causal Representation Learning for Multi-Armed Bandits with General Utilities},
  author={Wang, Ziniu and Li, Xiaotian and Zhao, Wenzhe and Xu, Qiongkai and Du, Yali and Zhan, Xianyuan and Wu, Yi-Fu and Zhang, Xiaoping and Zhou, Guyue and Zhu, Song-Chun},
  journal={arXiv preprint arXiv:2306.00836},
  year={2023}
}

@article{kunzel2019metalearners,
  title={Metalearners for estimating heterogeneous treatment effects using machine learning},
  author={K{\"u}nzel, S{\"o}ren R and Sekhon, Jasjeet S and Bickel, Peter J and Yu, Bin},
  journal={Proceedings of the National Academy of Sciences},
  volume={116},
  number={10},
  pages={4156--4165},
  year={2019},
  publisher={National Academy of Sciences}
}

@article{nie2020quasi,
  title={Quasi-oracle estimation of heterogeneous treatment effects},
  author={Nie, Xinkun and Wager, Stefan},
  journal={Biometrika},
  volume={108},
  number={2},
  pages={299--319},
  year={2020},
  publisher={Oxford University Press}
}

@article{kennedy2020optimal,
  title={Optimal doubly robust estimation of heterogeneous causal effects},
  author={Kennedy, Edward H},
  journal={arXiv preprint arXiv:2004.14497},
  year={2020}
}

@book{van2011targeted,
  title={Targeted learning: Causal inference for observational and experimental data},
  author={van der Laan, Mark J and Rose, Sherri},
  year={2011},
  publisher={Springer Science \& Business Media}
}

@article{carranza2022flexible,
  title={Flexible and Efficient Contextual Bandits with Heterogeneous Treatment Effect Oracles},
  author={Carranza, Aldo Gael and Krishnamurthy, Sanath Kumar and Athey, Susan},
  journal={arXiv preprint arXiv:2203.16668},
  year={2022}
}


\appendix
\section*{Appendix} 

\subsection{HDD Score Derivation}
\label{sec:appendix_hdd}
The HDD Score is calculated as the sample weighted sum of contextual KL divergence offset by the non-contextual KL divergence:
\begin{align*}
FI_{HDD}(x|m) &= \sum_{b=1}^{m_x} \frac{N_b}{N} D_b(P_{b_1}, ..., P_{b_k}) - D(P_{1}, ..., P_{k}) \\
&= \sum_{b=1}^{m_x} \frac{N_b}{N} \sum_{i=1}^k \sum_{j=1}^k \frac{N_{b_i} N_{b_j}}{N_b^2} D(P_{b_i}, P_{b_j}) \\
&\quad -  \sum_{i=1}^k \sum_{j=1}^k \frac{N_{i} N_{j}}{N^2} D(P_{i}, P_{j})\\
&= \sum_{b=1}^{m_x} \frac{N_b}{N} \sum_{i=1}^k \sum_{j=1}^k \frac{N_{b_i} N_{b_j}}{N_b^2} \sum_{v=0}^1 P_{b_i}(Y=v) \log \frac{P_{b_i}(Y=v)}{P_{b_j}(Y=v)} \\
&\quad - \sum_{i=1}^k \sum_{j=1}^k \frac{N_{i} N_{j}}{N^2} \sum_{v=0}^1 P_{i}(Y=v) \log \frac{P_{i}(Y=v)}{P_{j}(Y=v)}
\end{align*}
where $P_{b_i}(Y=v)$ is the probability of outcome $v$ for arm $i$ in bin $b$, and $N_{b_i}$ is the number of samples for arm $i$ in bin $b$. The KL divergence $D(P_X, P_Z)$ between two discrete distributions $P_X$ and $P_Z$ over outcomes $\{0,1\}$ is $\sum_{v=0}^1 P_X(Y=v) \log (P_X(Y=v)/P_Z(Y=v))$. The term $D_b(P_{b_1}, ..., P_{b_k})$ represents an average pairwise KL divergence between all arm distributions within bin $b$, and $D(P_{1}, ..., P_{k})$ is its global counterpart.

\subsection{Proofs}
\label{appendix:HIE-proof}

\subsubsection{Proof of Proposition~\ref{prop:HIE-nonnegative}}
\label{proof:HIE-nonnegative}

\begin{proof}
Let $w^*$ be the global winning arm and let $w_b$ be the best arm (winning arm) in bin $b$.
Recall that the HIE score is defined as:
\[
FI_{\text{HIE}}(x \mid m) 
= \sum_{b=1}^{m} \frac{N_b}{N} 
\Bigl[ 
    \max_{i \in \{1,\dots,k\}} P_{b_i}(1) 
    - \max_{i \in \{1,\dots,k\}} P_{i}(1)
\Bigr].
\]
Because
\[
\max_{i \in \{1,\dots,k\}} P_{b_i}(1) 
\; \ge \;
P_{b_{w^*}}(1) \quad \text{(since } w^* \text{ is one of the arms } i),
\]
and
\[
\max_{i \in \{1,\dots,k\}} P_{i}(1) 
= P_{w^*}(1)
= \sum_{b=1}^m \frac{N_b}{N}\, P_{b_{w^*}}(1) \quad \text{(by law of total probability)},
\]
it follows that
\[
\sum_{b=1}^m \frac{N_b}{N}\,\max_{i \in \{1,\dots,k\}} P_{b_i}(1)
\;\;\ge\;\;
\sum_{b=1}^m \frac{N_b}{N}\,P_{b_{w^*}}(1)
= \max_{i \in \{1,\dots,k\}}P_{i}(1).
\]
Subtracting 
$\max_{i \in \{1,\dots,k\}} P_{i}(1)$ from both sides (which is equivalent to subtracting $\sum_{b=1}^m \frac{N_b}{N} \max_{i \in \{1,\dots,k\}} P_{i}(1)$ from the summed term) yields 
\[
FI_{\text{HIE}}(x \mid m) 
= \sum_{b=1}^{m} \frac{N_b}{N} \max_{i \in \{1,\dots,k\}} P_{b_i}(1) - \max_{i \in \{1,\dots,k\}} P_{i}(1)
\; \ge \; 0.
\]
\end{proof}

\subsubsection{Proof of Proposition~\ref{prop:HIE-bias}}
\label{proof:HIE-bias}
\begin{proof}
Recall that for a contextual feature $x$ split into $m$ bins, the HIE score is given by
\[
FI_{HIE}(x \mid m) 
\;=\;
\sum_{b=1}^{m} \frac{N_b}{N} \max_{i \in \{1,\dots,k\}} P_{b_i}(1) \;-\;\max_{i \in \{1,\dots,k\}} P_{i}(1),
\]
where $N_b$ is the sample size in bin $b$, $N$ is the total sample size, 
$P_{b_i}(1)$ is the empirical probability of reward being $1$ for arm $i$ in bin $b$, 
and $\max_{i \in \{1,\dots,k\}} P_i(1)$ denotes the global best arm’s overall success probability (unconditional on $x$).
The function $f(p_1, \dots, p_k) = \max(p_1, \dots, p_k)$ is convex. By Jensen's inequality for expectations, if a bin $b$ is split into sub-bins $b_j$ with weights $w_j = N_{b_j}/N_b$ such that $\sum w_j = 1$, then $\mathbb{E}[\max_i P_{b_{j,i}}(1)] \ge \max_i \mathbb{E}[P_{b_{j,i}}(1)] = \max_i P_{b_i}(1)$, where the expectation is over the randomness of assigning samples to sub-bins if the split is finer than the true underlying data generation process for $P_{b_i}(1)$. More directly, consider the sum $\sum_{b=1}^{m} \frac{N_b}{N} \max_{i} P_{b_i}(1)$.
Let one bin $b_0$ be split into two sub-bins $b_{01}$ and $b_{02}$, with $N_{b_0} = N_{b_{01}} + N_{b_{02}}$.
The contribution to the sum from $b_0$ is $\frac{N_{b_0}}{N} \max_i P_{b_{0i}}(1)$.
After splitting, the contribution from $b_{01}$ and $b_{02}$ is $\frac{N_{b_{01}}}{N} \max_i P_{b_{01i}}(1) + \frac{N_{b_{02}}}{N} \max_i P_{b_{02i}}(1)$.
Since $P_{b_{0i}}(1) = \frac{N_{b_{01}}}{N_{b_0}} P_{b_{01i}}(1) + \frac{N_{b_{02}}}{N_{b_0}} P_{b_{02i}}(1)$, and $\max$ is a convex function, by Jensen's inequality:
\[
\frac{N_{b_{01}}}{N_{b_0}} \max_i P_{b_{01i}}(1) + \frac{N_{b_{02}}}{N_{b_0}} \max_i P_{b_{02i}}(1) \ge \max_i \left( \frac{N_{b_{01}}}{N_{b_0}} P_{b_{01i}}(1) + \frac{N_{b_{02}}}{N_{b_0}} P_{b_{02i}}(1) \right) = \max_i P_{b_{0i}}(1).
\]
Multiplying by $N_{b_0}/N$, we get:
\[
\frac{N_{b_{01}}}{N} \max_i P_{b_{01i}}(1) + \frac{N_{b_{02}}}{N} \max_i P_{b_{02i}}(1) \ge \frac{N_{b_0}}{N} \max_i P_{b_{0i}}(1).
\]
Thus, splitting a bin (or multiple bins) can only increase or maintain the value of the first term $\sum_{b=1}^{m} \frac{N_b}{N} \max_{i} P_{b_i}(1)$. Since the second term $\max_{i} P_{i}(1)$ is constant with respect to binning choices for feature $x$, the $FI_{HIE}(x|m)$ is non-decreasing as $m$ increases by splitting existing bins.
The same logic applies to the expected values:
\[
\mathbb{E}[FI_{HIE}(x \mid m + i)] 
\;\ge\; 
\mathbb{E}[FI_{HIE}(x \mid m)].
\]
\end{proof}

\subsection{Additional Figures}

\begin{figure*}[h] 
\centering
\includegraphics[width=0.6\linewidth]{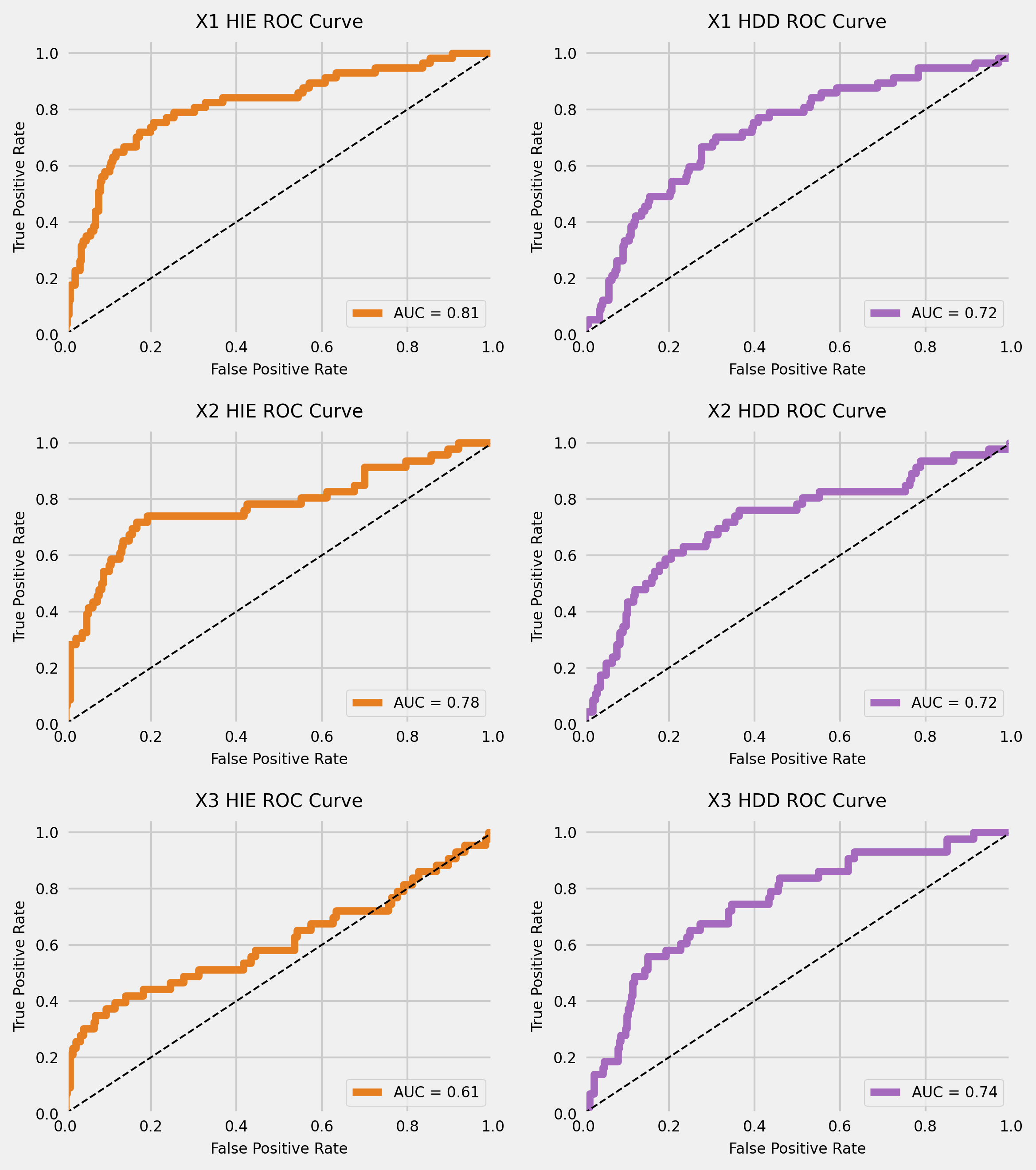} 
\caption{ROC curves for CMAB treatment effect significance, based on feature importance score p-values (feature scores are used to resolve ties).}
\label{fig:real_data_roc}
\end{figure*}

\subsection{Additional Tables}
\begin{table*}[htbp] 
\centering
\caption{Aggregated Evaluation Summary Across 10 Trials (Mean \textpm{} Std). Best scores per metric and sample size are in bold. HIE/HDD use $n_{bins}=20, p<0.1$. P@6/R@6 denote Precision/Recall at 6 features.}
\setlength{\tabcolsep}{4pt} 
\begin{tabular}{@{}ll S[table-format=1.3,table-figures-uncertainty=1]
                           S[table-format=1.3,table-figures-uncertainty=1]
                           S[table-format=1.3,table-figures-uncertainty=1]
                           S[table-format=1.3,table-figures-uncertainty=1]
                           S[table-format=1.3,table-figures-uncertainty=1]
                           S[table-format=1.3,table-figures-uncertainty=1]@{}}
\toprule
\multirow{2}{*}{Samples} & \multirow{2}{*}{Method} & \multicolumn{2}{c}{AUC-PR} & \multicolumn{2}{c}{Precision@6} & \multicolumn{2}{c}{Recall@6} \\
\cmidrule(lr){3-4} \cmidrule(lr){5-6} \cmidrule(lr){7-8}
 & & {Mean} & {Std} & {Mean} & {Std} & {Mean} & {Std} \\
\midrule
\multirow{8}{*}{1,000}   & HDD                     & \bfseries 0.813 & 0.112 & 0.433 & 0.179 & 0.433 & 0.179 \\
                         & HIE                     & 0.782 & 0.146 & 0.417 & 0.212 & 0.417 & 0.212 \\
                         & Pearson Correlation (abs) & 0.441 & 0.047 & 0.517 & 0.146 & 0.517 & 0.146 \\
                         & Mutual Information      & 0.374 & 0.041 & 0.367 & 0.105 & 0.367 & 0.105 \\
                         & Random Forest Imp.      & 0.362 & 0.019 & 0.317 & 0.095 & 0.317 & 0.095 \\
                         & CohortMAB Reward        & 0.623 & 0.181 & \bfseries 0.533 & 0.189 & \bfseries 0.533 & 0.189 \\
                         & LinUCB Reward           & 0.534 & 0.145 & 0.483 & 0.146 & 0.483 & 0.146 \\
                         & NonLinear LinUCB Reward & 0.632 & 0.159 & 0.583 & 0.196 & 0.583 & 0.196 \\
\addlinespace
\multirow{8}{*}{5,000}   & HDD                     & \bfseries 0.969 & 0.032 & \bfseries 0.900 & 0.086 & \bfseries 0.900 & 0.086 \\
                         & HIE                     & 0.927 & 0.035 & 0.783 & 0.081 & 0.783 & 0.081 \\
                         & Pearson Correlation (abs) & 0.475 & 0.018 & 0.650 & 0.053 & 0.650 & 0.053 \\
                         & Mutual Information      & 0.348 & 0.032 & 0.300 & 0.131 & 0.300 & 0.131 \\
                         & Random Forest Imp.      & 0.375 & 0.016 & 0.333 & 0.000 & 0.333 & 0.000 \\
                         & CohortMAB Reward        & 0.720 & 0.134 & 0.583 & 0.196 & 0.583 & 0.196 \\
                         & LinUCB Reward           & 0.808 & 0.089 & 0.700 & 0.153 & 0.700 & 0.153 \\
                         & NonLinear LinUCB Reward & 0.840 & 0.105 & 0.733 & 0.086 & 0.733 & 0.086 \\
\addlinespace
\multirow{8}{*}{10,000}  & HDD                     & \bfseries 0.997 & 0.008 & \bfseries 0.983 & 0.053 & \bfseries 0.983 & 0.053 \\
                         & HIE                     & 0.953 & 0.044 & 0.867 & 0.105 & 0.867 & 0.105 \\
                         & Pearson Correlation (abs) & 0.477 & 0.036 & 0.650 & 0.053 & 0.650 & 0.053 \\
                         & Mutual Information      & 0.350 & 0.023 & 0.283 & 0.081 & 0.283 & 0.081 \\
                         & Random Forest Imp.      & 0.380 & 0.012 & 0.333 & 0.000 & 0.333 & 0.000 \\
                         & CohortMAB Reward        & 0.872 & 0.067 & 0.767 & 0.117 & 0.767 & 0.117 \\
                         & LinUCB Reward           & 0.921 & 0.056 & 0.833 & 0.111 & 0.833 & 0.111 \\
                         & NonLinear LinUCB Reward & 0.868 & 0.083 & 0.750 & 0.118 & 0.750 & 0.118 \\
\addlinespace
\multirow{8}{*}{50,000}  & HDD                     & \bfseries 1.000 & 0.000 & \bfseries 1.000 & 0.000 & \bfseries 1.000 & 0.000 \\
                         & HIE                     & 0.951 & 0.024 & 0.850 & 0.053 & 0.850 & 0.053 \\
                         & Pearson Correlation (abs) & 0.480 & 0.017 & 0.667 & 0.000 & 0.667 & 0.000 \\
                         & Mutual Information      & 0.357 & 0.018 & 0.333 & 0.000 & 0.333 & 0.000 \\
                         & Random Forest Imp.      & 0.386 & 0.000 & 0.333 & 0.000 & 0.333 & 0.000 \\
                         & CohortMAB Reward        & 0.935 & 0.042 & 0.850 & 0.095 & 0.850 & 0.095 \\
                         & LinUCB Reward           & 0.959 & 0.045 & 0.867 & 0.105 & 0.867 & 0.105 \\
                         & NonLinear LinUCB Reward & 0.926 & 0.041 & 0.833 & 0.079 & 0.833 & 0.079 \\
\addlinespace
\multirow{8}{*}{100,000} & HDD                     & \bfseries 1.000 & 0.000 & \bfseries 1.000 & 0.000 & \bfseries 1.000 & 0.000 \\
                         & HIE                     & 0.954 & 0.031 & 0.867 & 0.070 & 0.867 & 0.070 \\
                         & Pearson Correlation (abs) & 0.473 & 0.028 & 0.667 & 0.000 & 0.667 & 0.000 \\
                         & Mutual Information      & 0.355 & 0.027 & 0.300 & 0.070 & 0.300 & 0.070 \\
                         & Random Forest Imp.      & 0.386 & 0.000 & 0.333 & 0.000 & 0.333 & 0.000 \\
                         & CohortMAB Reward        & 0.942 & 0.035 & 0.850 & 0.053 & 0.850 & 0.053 \\
                         & LinUCB Reward           & 0.971 & 0.026 & 0.900 & 0.086 & 0.900 & 0.086 \\
                         & NonLinear LinUCB Reward & 0.961 & 0.032 & 0.867 & 0.105 & 0.867 & 0.105 \\
\bottomrule
\end{tabular}
\end{table*}

\end{document}